\documentclass[runningheads]{llncs}
\usepackage{graphicx}

\begin{document}
\title{Light Communication for Controlling Industrial Robots}
%
%
%
\author{Fadi Al-Turjman\inst{1,2} \and
Diletta Cacciagrano\inst{3} \and
Leonardo Mostarda\inst{3} \and
Mattia Paccamiccio\inst{3} \and
Zaib Ullah\inst{3}
}
\authorrunning{Fadi Al-Turjman et al.}
%
%
\institute{Artificial Intelligence Department, Near East University, Nicosia, Mersin 10, Turkey \and
Research Center for AI and IoT, Near East University, Nicosia, Mersin 10, Turkey
\email{fadi.alturjman@neu.edu.tr}\\
 \and
Computer Science Division, University of Camerino, 62032 Camerino, Italy\\
\email{\{diletta.cacciagrano,leonardo.mostarda}
\email{mattia.paccamiccio,zaibullah.zaibullah\}@unicam.it}}

\maketitle              
%
\begin{abstract}
Optical Wireless Communication (OWC) is regarded as an auspicious communication approach that can outperform the existing wireless technology.
It utilizes LED lights, whose subtle variation in radiant intensity generate a binary data stream. This is perceived by a photodiode, that converts it to electric signals for further interpretation. This article aims at exploring the use of this emerging technology in order to control wirelessly industrial robots, overcoming the need for wires, especially in environments where radio waves are not working due to environmental factors or not allowed for safety reasons. We performed experiments to ensure the suitability and efficiency of OWC based technology for the aforementioned scope and "in vitro" tests in various Line-of-Sight (LoS) and Non-Line-of-Sight (NLoS) configurations to observe the system throughput and reliability. The technology performance in the "clear LoS" and in the presence of a transparent barrier, were also analyzed.

\keywords{Visible Light Communication \and Optical Wireless Communication \and Industrial Robots \and Performance.}
\end{abstract}
%
%
%

\section{Introduction}\label{intro}
Autonomous robots are a crucial component of Industry 4.0 \cite{Lu2017}. They can be used in order to improve the speed and accuracy of operations, especially in warehousing and manufacturing environments. Robots can work alongside with humans for added efficiency while reducing the employee injury risk in dangerous environments. To date most of the commercially available robots are supported by wired consoles that are used to move the robot during its programming stage while in production its motion is automated. The necessary features that a robot console should have are joint movements, dead-man button, and emergency stop. The dead-man button is a switch that is activated or deactivated if the human operator becomes unable, such as through death, loss of consciousness, or being bodily removed from control. The robot is allowed to move until such button is pressed, once it is released the robot is prevented to move, both by stopping the motions and by preventing new motions to start. The emergency stop button is a mushroom-headed red button that, when pressed, will immediately stop the robot. This is typically implemented in a purely hardware fashion in order to have a high reliability. The emergency stop uses a communication line that is separated from the robot manoeuvring communication line. This is done to totally exclude software logic that could impact on the reliability and real-timeness of this critical feature. Although wired consoles can ensure real-time and reliable communication they reduce the operator degree of movement hence efficiency. This is way various wireless console by using Radio Frequency (RF) based solutions have been attempted (e.g., WiFi and Bluetooth). The problem of these solutions is that they fail to operate in many real industrial factories. This can be consequence of harsh signal propagation conditions together with interference with coexisting radio technologies that operate in the same frequency. This may lead to poor network performance or even  failures \cite{Wetzker2016}. The contribution of this paper is the application of the OWC technology for wireless manoeuvring of robots in industrial environment. 

\subsection{Optical Wireless Communication at glance}
OWC is a communication standard that operates in the electromagnetic spectrum of light, which is 1mm to 10 nm wavelength. A photodiode acts as a receiver while a light source performs as a data transmitter \cite{martinek2019visible}, \cite{biton2018improved}. OWC can enhance the data rate capacity of wireless networks, enables energy-efficient communication, and materializes data communication in susceptible environments.  OWC is certain to replace the older radio waves based technology due to the highly congested radio spectrum, at least in very localized environments such as supermarkets, offices and industries. Currently various research challenges such as modulation schemes, throughput and advanced networking are being investigated to achieve the highest data rates for next-generation communication \cite{shao2015design}. Many OWC-based applications have been developed, including toys, air conditioners, TV remotes controllers, human sensing, vehicle-to-vehicle communication, underwater communication and bar code readers. In the near future, unmanned aerial vehicles (UAVs) will play a significant role in commercial activities and are expected to make use of OWC technologies besides radio frequency technologies for stable, secure and high bandwidth communications \cite{soner2019low,ullah2020cognition}. The academia and research industry have established the IEEE 802.11bb task group to modify the existing MAC and physical layer according to OWC \cite{galisteo2019research}.

\section{Paper contribution}
The contribution of this paper is the application of the OWC technology for wireless manoeuvring of robots in industrial environment. To this ending we need to address the following basic questions: (i) would OWC have satisfactory performance in terms of latency, reliability and throughput when compared to wired and RF-based wireless communication?; (ii) will these non functional requirements be satisfied in case of a moving operator?; (iii) how can we implement the dead-man button and emergency stop channel with the highest possible reliability?. In order to answer these questions we performed cautious 'in vitro' experiments to test the latency, reliability and throughput of a prototype OWC system. After successful completion of the 'in vitro' settings we replicated the experiments in real case study scenarios for moving robots inside various farms. Experiments were performed in various LoS and NLoS configurations to observe the system throughput and reliability. The technology performance in the "clear LoS" and in the presence of a transparent barrier, were also analyzed. We also analyzed the use of a pointing system for the link to be kept always in optimal conditions and proposed an additional communication channel for a reliable emergency stop. Our experiments show that OWC based communication can be used for wireless manoeuvring of robots.

This article can be summarised as follows: Section \ref{Literature} presents a review of the most recent and relevant literature, Section \ref{Model} presents our experimental setup, Section \ref{benchmark} provides a detailed analysis of the experimental results and Section \ref{conclusion} concludes the article and offers future research directions. 
    
\section{Literature Review}\label{Literature}
In this section we report various case studies where OWC technology has been tested. 

\textbf{OWC based  vehicle  collision  avoiding  system}
Intelligent transportation system and vehicles safety are one of primary concerns of smart city projects \cite{ullah2020applications,8732936,8624288}. The researchers in academia and industry are working to develop advanced systems that could drastically minimize the rate of accidents \cite{masini2018survey,goto2016new,raza2018social}.
In order to achieve the preceding goal, OWC has been proposed as a possible solution because of its properties. First of all it has a huge and unregulated spectrum, high transmission capacity, and mature enough LED technology. Many solutions have been proposed, such as microwave radar systems and short range radio systems, but all of these architectures suffer from frequency competition and weather conditions changing. These studies focus on the rear-end collision scene, that is the most common type of accident.
    	
In \cite{siddiqi2016visible}, the authors have proposed a OWC based prototype to enhance the safety and efficiency of intelligent transport systems (ITS). In the proposed methodology, light emitted from the brake lamps of a vehicle can be used to transmit messages to the following vehicle so that necessary safety measures can be taken in time. The experimental results show that the prior prototype can identify hard brakes over a distance of 20 meters and can offer an alert warning to following vehicles driving slower than 80 km/h.
The authors in \cite{bao2016visible} have also proposed a OWC based rear-collision avoidance system. The efficiency of the proposed technique has been verified by implementing different case studies.
In \cite{Dahri}, the authors developed a OWC-based V2V communication prototype. Various techniques like frequency shift keying (FSK), phase shift keying, and amplitude shift keying were implemented. Different measuring parameters e.g., received optical power, bit error rate (Ber), and received signal voltage were assessed to characterize V2V communication. The results reveals that 3.5 Mb/s and 500 Kb/s data rates were achieved over the distances of 0.5 and 15 meters respectively. The experimental results validated the efficiency of OWC-based proposed architecture and its importance in V2V communication.     
    	
\textbf{OWC and Underwater Robots}
\label{subsec:underwaternuclear}
Nowadays, OWC is emerging as a method to provide high data rates and low latency for underwater wireless communication \cite{zhou2019common}. The OWC-based systems outperform radio wireless in terms of reach and bandwidth with the LoS as a prerequisite. The proposed solution to this problem is to implement a feedback control to direct the light emitters and detectors to each other.

The system is motivated by the need for wireless communication systems in the robotic inspection of nuclear reactors, which are permanently underwater. An approach based on OWC has been pursued due to the significant range and bandwidth advantages of the technology in this specific environment.

There is an urgent need for a more thorough investigation of underwater structures as an added safety measure. The need for a pointing system is crucial as the optical components are inherently directional and require a continuous LoS to maintain a data link. 

The authors in \cite{rust2012dual} mainly focus on the use of light as a localization medium via an unusual general strategy, where the light source is integrated into a full inertial measurement unit for estimation of orientation and position. In this way, the optical communication system is in a dual-use configuration. The light signal is both interpreted using a circuit to attain a data signal and analyzed using different sensors and techniques to gain an estimate of the orientation and position of the vehicle.

\textbf{OWC Technology based Robots in Pipelines}
\label{subsec:pipelinerobots}
In the near future, robots are expected to be employed in pipelines with protracted distances and complicated networks. These robots need to be highly efficient in terms of wireless communication but the presence of EM interference, Faraday cage effects and low energy efficiency can undermine their performance in said environment.
In \cite{zhao2019preliminary}, the authors proposed to employ a wheeled robot equipped with a OWC-based transmitter and receiver to assess their efficacy inside a pipeline. The attained experimental results are satisfactory as the proposed system is capable to establish a good communication link and provide illumination inside the pipe.

\textbf{ OWC for Enhanced Control of Autonomous Delivery Robots}
\label{subsec:deliveryrobots}
The existing robot systems are mainly focused on robot functionality and position accuracy with fewer concerns about their safety. Regarding robot safety, OWC shows promising potential. The authors in \cite{murai2012novel}, developed a robot having an enhanced navigation control system using a joint of navigation sensors and OWC-based technology using in-building installed LEDs. The proposed robot system has been tested in a real hospital and shown satisfactory experimental results.   

\section{Experimental setup}\label{Model}
In this section we explain all details that are  related to our experiments. We provide hardware, software and strategies chosen together with the motivations that led to such choices. 

\begin{figure}[!tbp]
  \centering
  \begin{minipage}[b]{0.4\textwidth}
  \centering
    \includegraphics[width=\textwidth]{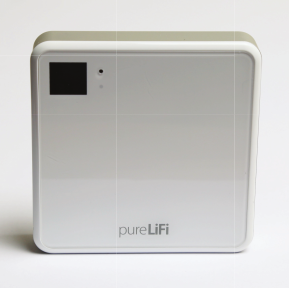}
    		\caption{The access point.}
    		\label{fig:Accesspoint}
  \end{minipage}
  \hfill
  \begin{minipage}[b]{0.45\textwidth}
  \centering
    \includegraphics[width=\textwidth]{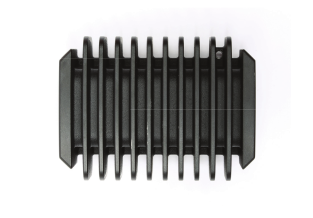}
    		\caption{The TX driver.}
    		\label{fig:drivers}
  \end{minipage}
\end{figure}

\begin{figure}[htpb!]
    		\centering
    		\includegraphics[width=0.2\textwidth]{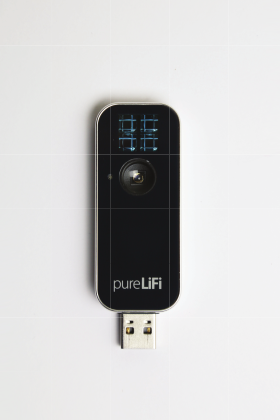}
    		\caption{The USB dongle.}
    		\label{fig:usbdongle}
    	\end{figure}

Our experiments were performed by using the development kit manufactured by pureLiFi. It consists of two USB Li-Fi dongles (see Figure \ref{fig:usbdongle}), two Li-Fi ready LED lamps and two access points (see Figure \ref{fig:Accesspoint}). The access point (in which is implemented an infrared receiver) is wired to the lamp via the TX driver (see Figure \ref{fig:drivers}) and to the rest of the wired network. It acts as a signal modulator and is responsible for generating the signal the lamp will reproduce. 

The dongle is wired to the computer via USB and it has implemented an infrared transmitter and a visible light photodiode so that it can transmit and receive data.
    	
The data link layer of the pureLiFi system is compliant with the 802.11 protocol (CSMA/CA with RTS/CTS and ACK) while its physical layer uses light has the medium of data communication.

\subsection{Reference system architecture}

\begin{figure}[htpb!]
		\centering
		\includegraphics[width=0.78\textwidth]{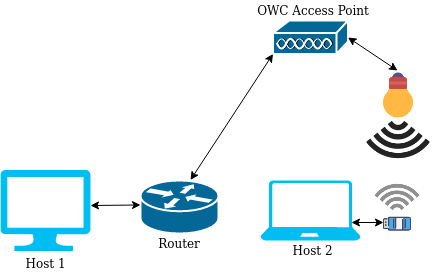}
		\caption{The proposed architecture for test setup.}
		\label{fig:lifiarchitecture}
	\end{figure}

\begin{figure}[!tbp]
  \centering
  \begin{minipage}[b]{0.4\textwidth}
  \centering
		\includegraphics[width=\textwidth]{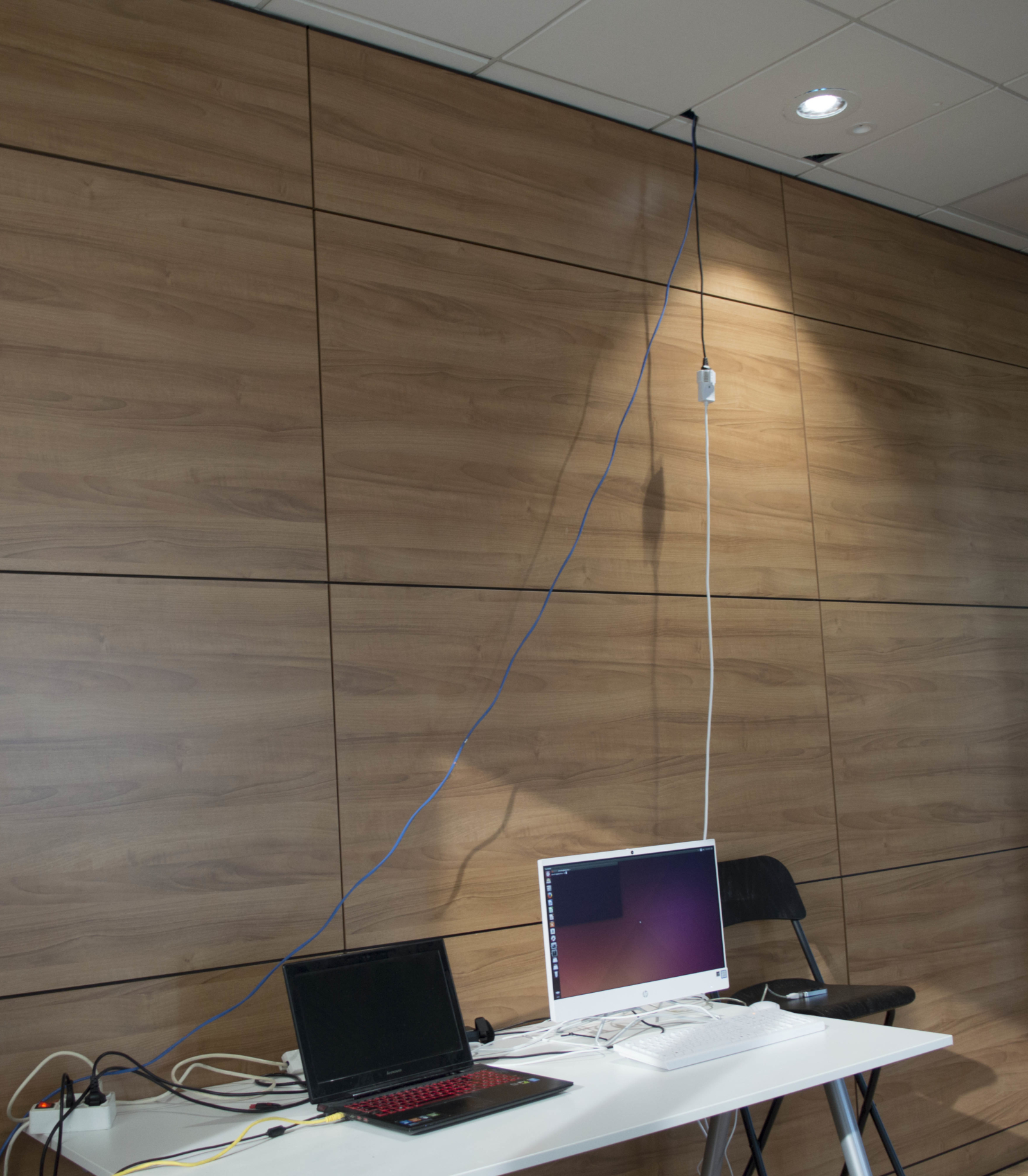}
		\caption{The "in vitro" environment.}
		\label{fig:envmisco}
  \end{minipage}
  \hfill
  \begin{minipage}[b]{0.45\textwidth}
  \centering
		\includegraphics[width=\textwidth]{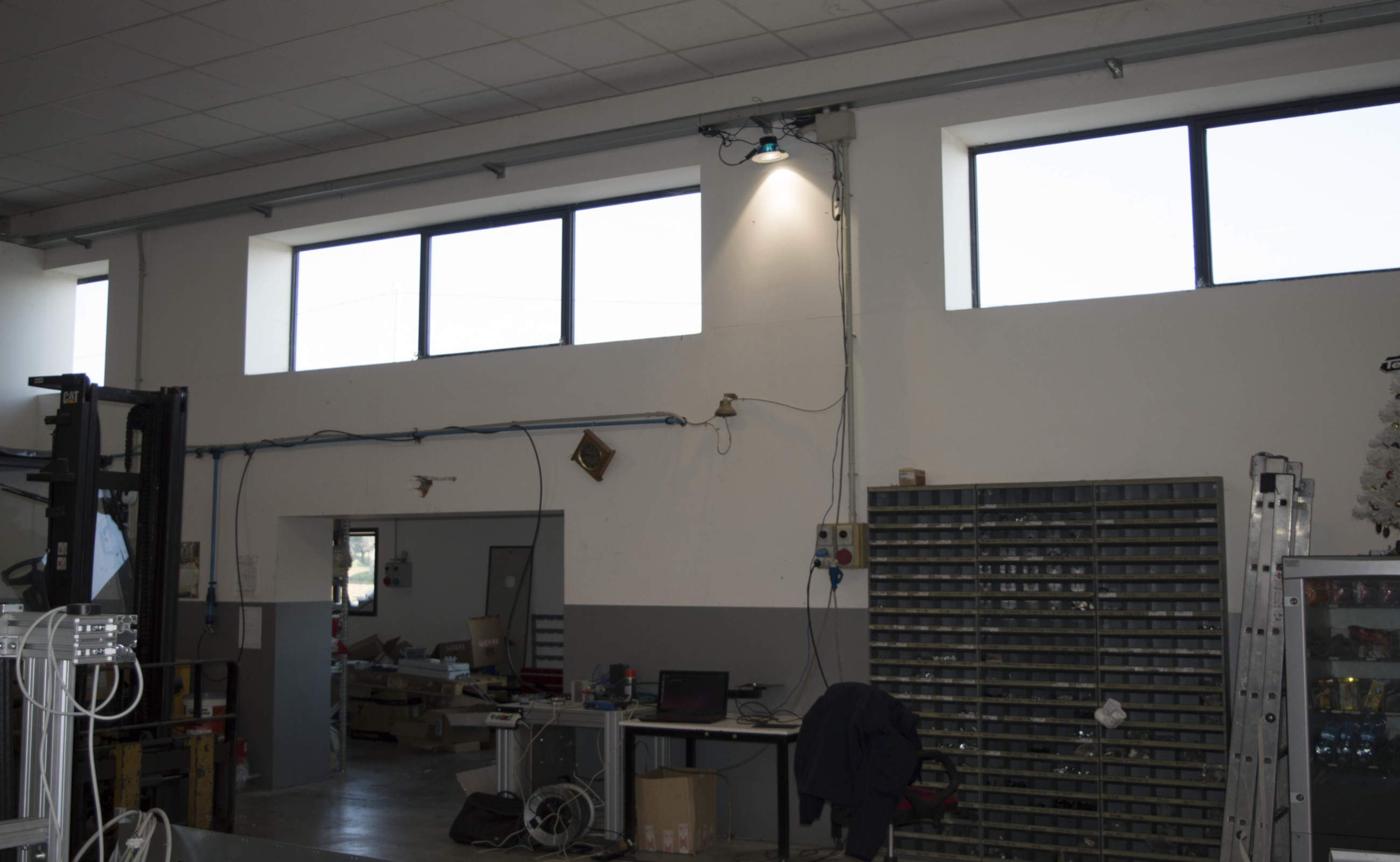}
		\caption{The "real life" environment.}
		\label{fig:envtekna}
  \end{minipage}
\end{figure}

Figure \ref{fig:lifiarchitecture} and \ref{fig:envmisco} show the setup we used for our 'in vitro' experiments. Communication was performed by assuming that a host computer 2 (this simulates the console) exchanges data with a host computer 1 (this simulates the robot's continuous numeric control (CNC)). The continuous numeric control provides instructions on where and if to move the robot's joints. The console is connected with the OWC access point while the CNC is connected with a local router. Router and access point are connected by means of a local Ethernet cable. We used this setting in order to emulate the robot controlling scenario.

In the 'in vitro' experiments the wired network was isolated from the traffic, i.e., only the traffic sent via OWC was observable and we made sure that no light interference was present. This was not the same in the industrial experiments where interference was possible.

\subsection{Performance measures}

Our aim is to estimate reliability in terms of Packet Error Rate (PER), throughput and Latency. These were evaluated by performing various experiments (e.g., LoS and NLoS).

\subsubsection{Packet error rate (PER) estimation.} 

We are interested in evaluating the PER that is related to the link between the console and the OWC access point. This can give an estimate on the quality of the light channel under various conditions. There are various methods for performing PER tests for wireless networks \cite{1429989}. On way is to compare the raw data bits that are received by the OWC access point with the ones that are sent by the consoles. This can allow us to measure the Bit Error Rate (BER) that can be used in order to estimate the PER. This is calculated by assuming a uniformly distributed error which can lead to gross overestimation of PER. Another way to estimate the PER is to count the number of CRC mismatches at the OWC access point. With the 32-bit CRC used by the 802.11 standard the probability of undetected erroneous packets is very small (i.e., 2.3E-10). Both methods (BER and CRC) require  a specific vendor software  to get data from the MAC layer. When this is not available, packets with an unreliable protocol (such as UDP) can be transmitted. The PER can be obtained by counting the number of missing packets at the router side since any packets with errors is dropped. We use this technique in order to estimate the quality of the connection between the OWC access point and the console \footnote{This measure gives a good indication of the OWC to console connection. In fact, the wired connection between router and OWC has a constant and very low PER.} since no specific vendor software to get OWC MAC layer data was available.

For each time slot $T_{i}$ (all time slots have equal duration) we have calculated the average packet error rate  $PER_{T_{i}}$ according to the following formula: 
	
$$PER_{T_{i}}=\frac{F^{A}_{T_{i}}-F^R_{T_{i}}}{F^{A}_{T_{i}}}$$
	    
where $F^{R}_{T_{i}}$ is the number of frames that were received at the router during transmission between the console and the router via OWC; $F^{A}_{T_{i}}$ is the total number of frames that were sent; $F^{A}_{T_{i}}-F^R_{T_{i}}$ is the number of frame that were not received.
	  
The average error rate $PER$ was calculated according to the following formula:
	    
$$PER=\frac{\sum_{i=0}^{N}PER_{T_{i}}}{N}$$
	
where $N$ is the total number of time slots.
	  
The confidence interval of the experiment was calculated in the following manner:
	    
$$PER \pm Z \frac{s}{\sqrt{N}}$$
	    
where $s$ is the standard deviation and $Z$ is the confidence interval (0.95 in our case) and $Z \frac{s}{\sqrt{N}}$ the margin of error. In all graphs we always plot the average in blue and we always plot the statistics endpoints $\overline{PER} - Z \frac{s}{\sqrt{N}}$ and   $\overline{PER} + Z \frac{s}{\sqrt{N}}$ in red and yellow, respectively. For instance, Figure \ref{fig:speedverticallosvitro} shows in blue the average PER, in red the lower endpoint and in blue the upper one. 
 
For each experiment, we have collected several days of data exchange. More precisely, for each day all packets exchanged during 4.5 hours of communication between console and OWC have been collected. These have been divided into chunks of 10 minutes. For each chunk we calculated $PER_{T_{i}}$. These have been  averaged together in order to obtain the PER of an experiment. 

We have implemented a client UDP program that sends packets to a server program. Sent and received packets have been counted at the client and server side, respectively. We have double checked the packet counting by using the Wireshark packet analyser.


\subsubsection{Throughput}
	 
Throughput is the number of messages that are successfully delivered per unit of time. This is consequence of the available bandwidth, quality of links (error rate) and hardware limitations. Throughput is measured from the arrival of the first bit from console at the router. This is done in order to decouple the concept of throughput from the concept of latency. We use a TCP throughput based estimation where for each time slot $T_{i}$ the console sends to the router packets as fast as the hardware will allow. We calculated the throughout $R_i$ for the time slot $T_{i}$ by using the following formula:

$$R_i=\frac{B_i}{T_{i}}$$

where $B_i$ is the total amount of byte that were sent during the time slot $T_i$. For each experiments we collected thousands of time slots $T_i$ which were used to calculate the following average throughput rate $R$:

$$R=\frac{\sum_{i=0}^{N}R_i}{N}$$

where $N$ is the total amount of time slots and $R_i$ is the throughout of the time slot $T_i$. The confidence interval of the experiment was calculated by using the following formula:
	    
$$R \pm Z \frac{s}{\sqrt{N}}$$
	    
where $s$ is the standard deviation and $Z$ is the confidence interval (0.95 in our case) and $Z \frac{s}{\sqrt{N}}$ the margin of error. Like for the PER, in all graphs we always plot the average in blue and we always plot the statistics endpoints $\overline{R} - Z \frac{s}{\sqrt{N}}$ and   $\overline{R} + Z \frac{s}{\sqrt{N}}$ in red and yellow, respectively. 

We have implemented a client TCP program that sends packets to a server program. The client runs on the console while the server at the router. The time to receive the packets have been taken at the client side. We have double checked the throughput results obtained by using the Wireshark packet analyser and the iPerf 3 tool. This is a widely used command-line tool written in C for network performance measurement. We did not use iPerf only since it might use other protocols such as TELNET or serial to output the intermediary results at each interval which might introduce undesired overhead. This may negatively impact the throughput results.

	 

\subsubsection{Latency}

We use the round-trip time (RTT) in order to estimate the time it takes for an acknowledgement to be received by the console. More precisely, we measure the time it takes for 512bytes to be sent by the console to the router and the related acknowledgement to be received back. We have chosen 512 byte since it is a sufficient amount of data for controlling the movement of the robots. 

For each slot $S_i$ we have estimated the average $RTT_{i}$ by applying the following formula:

$$RTT_{i}=\frac{\sum_{j=0}^{j=N_i}T(P_j)}{N_i}$$

where $T(P_j)$ is the round-trip time for the packet $P_j$ and $N_i$ the number of packet sent for the slot $S_i$ (this has been set to a thousand).We have also calculated the $peaks_{i}$. This counts the number of packet whose $RTT$ exceeds the time of $30ms$ for the slot $S_i$. Exceeding this time is considered no safe when controlling the arm of a robot. 

For each experiment we have calculated the average $RTT$ by using the following formula:

$$RTT=\frac{\sum_{j=0}^{j=N}RTT_{i}}{N}$$

where $N$ is the number of slots which have been set to a thousand. The total percentage of peaks was calculated according to the following formula:

$$peaks=\frac{\sum_{j=0}^{j=N}peaks_{i}}{\sum_{j=0}^{j=N}N_{i}}$$

where $\sum_{j=0}^{j=N}N_{i}$ is the total amount of packets sent for all slots. 

The confidence interval of the experiment was calculated in the following manner:
	    
$$\overline{RTT} \pm Z \frac{s}{\sqrt{N}}$$
	    
where $s$ is the standard deviation and $Z$ is the confidence interval (0.95 in our case) and $Z \frac{s}{\sqrt{N}}$ the margin of error. Like for the PER, in all graphs we always plot the average in blue and we always plot the statistics endpoints $\overline{RTT} - Z \frac{s}{\sqrt{N}}$ and   $\overline{RTT} + Z \frac{s}{\sqrt{N}}$ in red and yellow, respectively.

We have implemented a client TCP program that sends packets to a server program. The client runs on the console while the server at the router. The time to receive the ack on the client side has been recorded every time. We have double checked the throughput results obtained by using the Wireshark packet analyser and the TCP-latency tool. This is a command-line tool implemented in Python that is born from the need of running network diagnosis tasks on serverless infrastructure (many providers do not include ICMP support).




\section{Experiments and test Results}\label{benchmark}
In this section, we describe all experiments that we have performed and we discuss their results.

\subsection{Throughput and PER tests}
\label{ThroughputPER}
We performed various experiments in order to observe the variation in throughput, PER and Latency by changing the following settings: (i) LoS and NLoS; (ii) the vertical position of the lamp from 0.5m to 5m while keeping the dongle stationary under the lamp ( horizontal position 0m); (iii) changing the horizontal position of the dongle from 0m to 2m while keeping the vertical position of the lamp at 2.5m and 5m from the ground. In line-of-sight the coupled transmitters and receivers directly "face" each other while in NLoS the signal is not carried by the light beam directly but by its reflection.

\subsubsection{"In vitro" tests}

\begin{figure}[!tbp]
  \centering
  \begin{minipage}[b]{0.4\textwidth}
  \centering
	\includegraphics[width=1.3\textwidth]{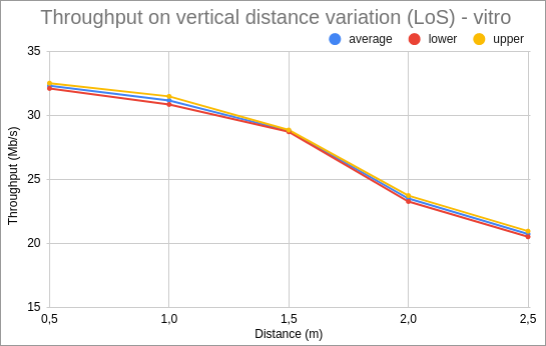}
	\caption{Throughput for vertical distance variation, LoS.}
	\label{fig:speedverticallosvitro}
  \end{minipage}
  \hfill
  \begin{minipage}[b]{0.45\textwidth}
  \centering
	\includegraphics[width=1.15\textwidth]{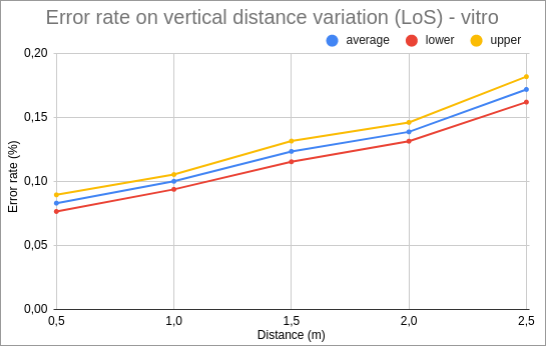}
	\caption{Error rate for vertical distance variation, LoS.}
	\label{fig:errorrateverticallosvitro}
  \end{minipage}
\end{figure}

\begin{figure}[!tbp]
  \centering
  \begin{minipage}[b]{0.4\textwidth}
 	\centering
	\includegraphics[width=1.4\textwidth]{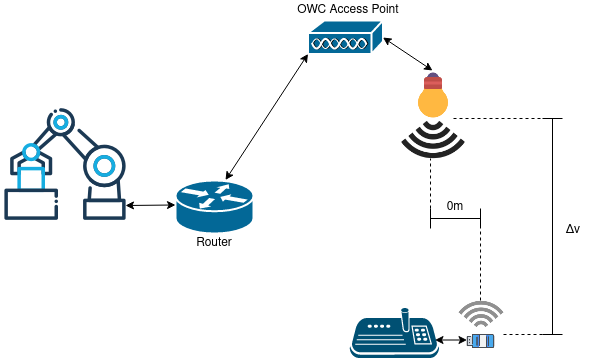}
	\caption{Vertical 'in vitro' scenario  with LoS.}
	\label{fig:vitro_los_vertical}
  \end{minipage}
  \hfill
  \begin{minipage}[b]{0.45\textwidth}
  	\centering
	\includegraphics[width=1.3\textwidth]{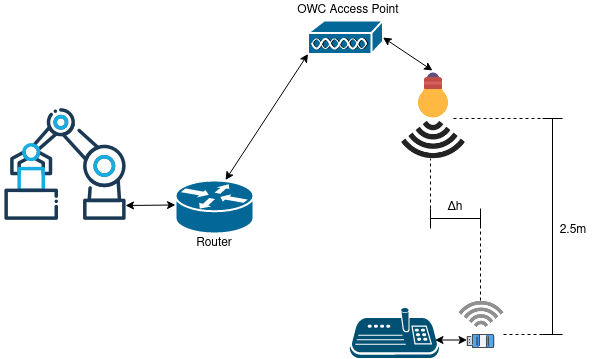}
	\caption{Horizontal 'in vitro' scenario  with LoS.}
	\label{fig:vitro_los_horizontal}
  \end{minipage}
\end{figure}

Figures \ref{fig:speedverticallosvitro} and \ref{fig:errorrateverticallosvitro} show the results for LoS tests where the dongle position varies vertically between 0.5 and 2.5m. This scenario is shown is Figure \ref{fig:vitro_los_vertical}. As expected, the speed and error rate have opposite behaviour. From the results, it is evident that as the distance increases, the throughput decreases. At the test's poles, the speed dropped by approximately one-third of the peak value and the error rate is doubled. 

\begin{figure}[!tbp]
  \centering
  \begin{minipage}[b]{0.4\textwidth}
  \centering
	\includegraphics[width=1.25\textwidth]{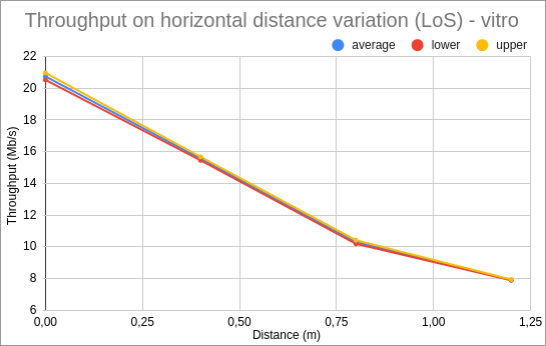}
	\caption{Throughput for horizontal distance variation, LoS.}
	\label{fig:speedhorizontallosvitro}
  \end{minipage}
  \hfill
  \begin{minipage}[b]{0.45\textwidth}
  \centering
	\includegraphics[width=1.1\textwidth]{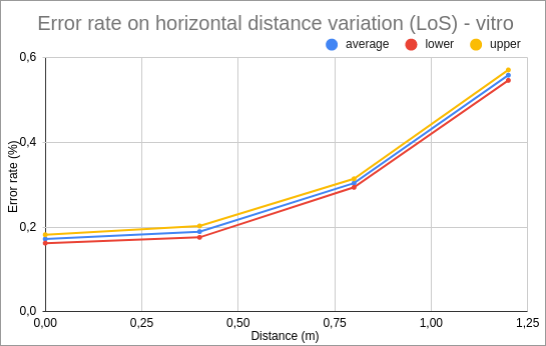}
	\caption{Error rate for horizontal distance variation, LoS.}
	\label{fig:errorratehorizontallosvitro}
  \end{minipage}
\end{figure}

Figures \ref{fig:speedhorizontallosvitro} and \ref{fig:errorratehorizontallosvitro} present the results for LoS tests where the dongle position varies horizontally (with a fixed vertical distance of 2.5 meters from the light source). More precisely, the horizontal distance varies between 0 and 1.25m from the centre of the cone light. This setting is shown in Figure \ref{fig:vitro_los_horizontal}. We can observe that the speed and error rate have diverging responses. At the extreme values of the test, speed dropped by around three times from the peak performances and the error rate gets tripled.

\begin{figure}[!tbp]
  \centering
  \begin{minipage}[b]{0.4\textwidth}
  \centering
	\includegraphics[width=1.2\textwidth]{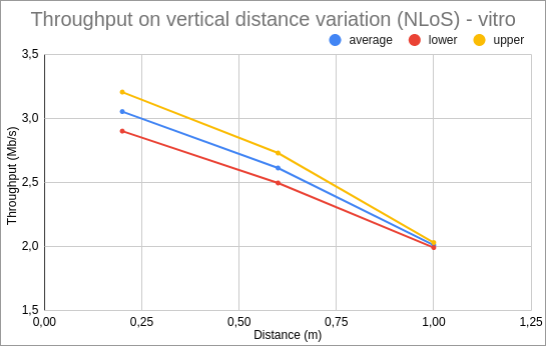}
	\caption{Throughput for vertical distance variation, NLoS.}
	\label{fig:speedverticalnlosvitro}
  \end{minipage}
  \hfill
  \begin{minipage}[b]{0.45\textwidth}
	\centering
	\includegraphics[width=1.1\textwidth]{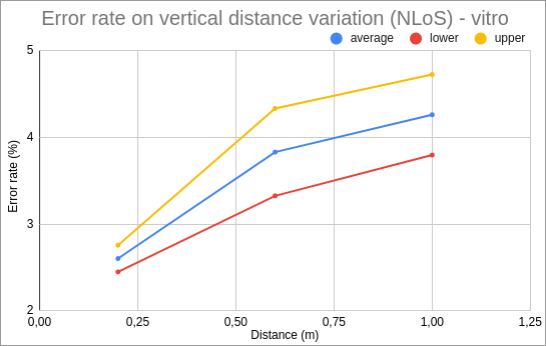}
	\caption{Error rate for vertical distance variation, NLoS.}
	\label{fig:errorrateverticalnlosvitro}
  \end{minipage}
\end{figure}

\begin{figure}[htpb!]
	\centering
	\includegraphics[width=0.6\textwidth]{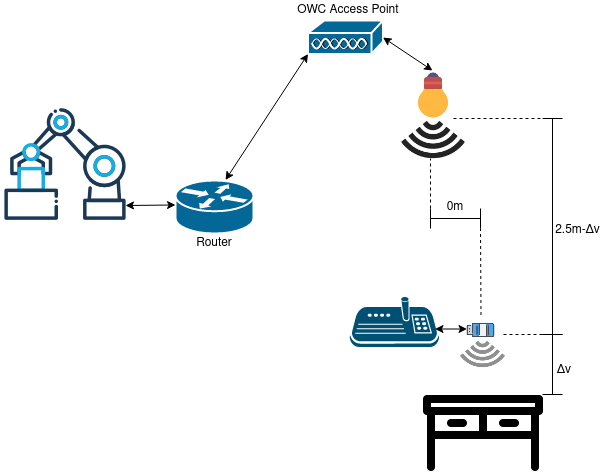}
	\caption{Vertical 'in vitro' scenario  with NLoS.}
	\label{fig:vitro_nlos_vertical}
\end{figure}

Figures \ref{fig:speedverticalnlosvitro} and \ref{fig:errorrateverticalnlosvitro} depict the results for NLoS tests where the position of the dongle varies vertically (the distance is considered from the surface reflecting light). Figure \ref{fig:vitro_nlos_vertical} shows this setting. We can see that the console is faced toward a reflecting surface. The distance from this surface ( $\Delta v$ in Figure \ref{fig:vitro_nlos_vertical}) is varied between 0.25m and 1m. Figures \ref{fig:speedverticalnlosvitro} and \ref{fig:errorrateverticalnlosvitro} show that at the polar values of the experiment, speed reduces by approximately one third from the peak value and the error rate almost doubled. Here the values in terms of throughput are much lower than the LoS configuration and the error rate is also higher.

\begin{figure}[!tbp]
  \centering
  \begin{minipage}[b]{0.4\textwidth}
  \centering
	\includegraphics[width=1.25\textwidth]{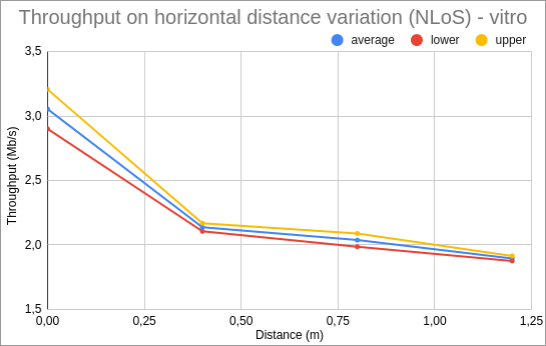}
	\caption{Throughput for horizontal distance variation, NLoS.}
	\label{fig:throughputhorizontalnlosvitro}
  \end{minipage}
  \hfill
  \begin{minipage}[b]{0.45\textwidth}
  \centering
	\includegraphics[width=1.1\textwidth]{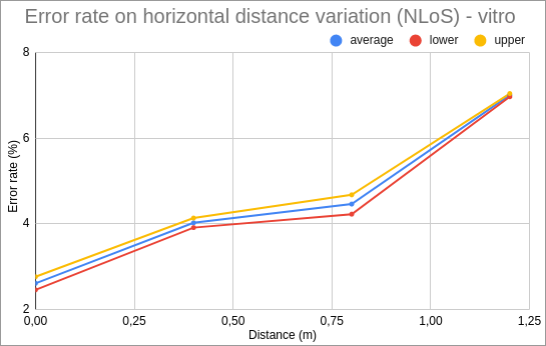}
	\caption{Error rate for horizontal distance variation, NLoS.}
	\label{fig:errorratehorizontalnlosvitro15}
  \end{minipage}
\end{figure}

Figures \ref{fig:throughputhorizontalnlosvitro} and \ref{fig:errorratehorizontalnlosvitro15} portray the results for the NLoS experiments where the position of the dongle changes horizontally (the vertical distance is constant and is 0.2 meters, the nearest possible to the reflected light source so that the reflection of the infrared light can reach the access point).
Considering the extreme values of the attained results (Figures \ref{fig:throughputhorizontalnlosvitro} and \ref{fig:errorratehorizontalnlosvitro15}), the speed becomes almost half of the peak value and the error rate exceeds the threefold.

\subsubsection{Experiments inside the farms}
\begin{figure}[!tbp]
  \centering
  \begin{minipage}[b]{0.4\textwidth}
  \centering
	\includegraphics[width=1.25\textwidth]{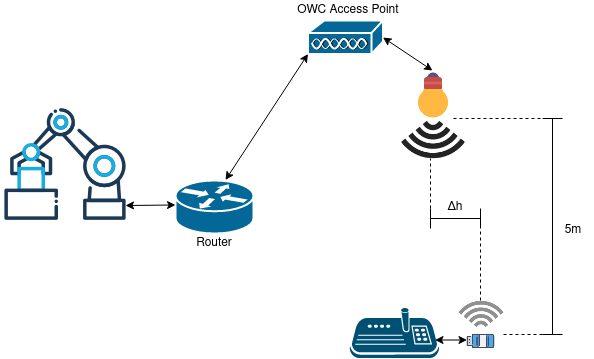}
	\caption{Experiments inside the industry with LoS.}
	\label{fig:reallife_clear}
  \end{minipage}
  \hfill
  \begin{minipage}[b]{0.45\textwidth}
  	\centering
	\includegraphics[width=1.1\textwidth]{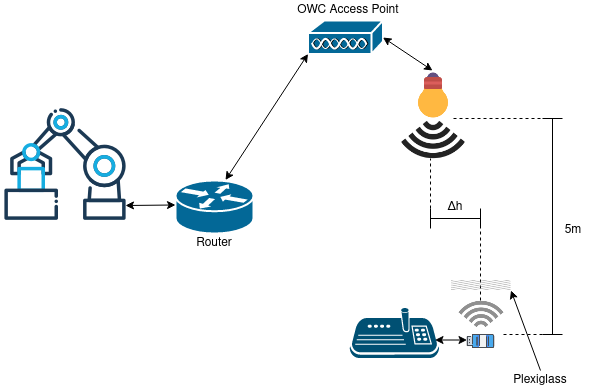}
	\caption{Experiments inside the industry with a transparent obstacle in the middle}
	\label{fig:reallife_plexi}
  \end{minipage}
\end{figure}

The first noticeable outcome of farm experiments was that the NLoS experiments inside all farms were not possible since communication links simply failed to establish. The distances involved were to large (see Figure \ref{fig:envtekna}) thus the reflection of the light source was too weak to be received at the OWC access point.
    
The experiments inside the farms were only performed by varying the horizontal distance from the center of the light cone. The operator's console would be around 1.5 meters from the ground in the best case while the lamp was about 5 meters from the console. This was the height of all the warehouse ceiling where the experiments were performed. Two types of experiments were conducted, in one case the console and the lamp were in LoS without any obstacle in between (see Figure \ref{fig:reallife_clear}). In the second case a transparent barrier during between the OWC communication and the console was placed (see Figure \ref{fig:reallife_plexi}). This was done in order to simulate a plexiglass material that could separate the operator and the robot.  The addition of the transparent barrier definitely affected the packet error rate and the throughput but real time (i.e., response time) within certain time limit can be still ensured.
    
\begin{figure}[!tbp]
  \centering
  \begin{minipage}[b]{0.4\textwidth}
  \centering
	\includegraphics[width=1.25\textwidth]{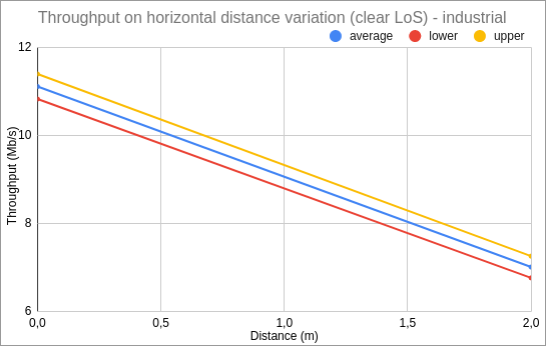}
	\caption{Throughput for horizontal distance variation in LoS test.}
	\label{fig:speedhorizontallosreallife}
  \end{minipage}
  \hfill
  \begin{minipage}[b]{0.45\textwidth}
  	\centering
	\includegraphics[width=1.1\textwidth]{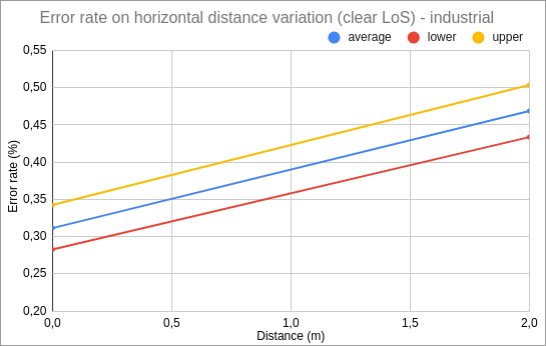}
	\caption{Error rate for horizontal distance variation in LoS test.}
	\label{fig:errorratehorizontallosreallife}
  \end{minipage}
\end{figure}

Figure \ref{fig:speedhorizontallosreallife} and \ref{fig:errorratehorizontallosreallife} show the throughput  and the packet error rate when the dongle position  varies  horizontally from 0 to 2m and no obstacle is used. It is worth noticing that the throughput almost halved when compared to the 'in vitro' experiments of 	Figures \ref{fig:speedhorizontallosvitro}. This is because the height of the lamp doubled (from 2.5m to 5m). It is also worth noticing that the packet error rate increased approximately by 50\%  when compared to the 'in vitro' experiments of Figure \ref{fig:errorratehorizontalnlosvitro15}. 

\begin{figure}[!tbp]
  \centering
  \begin{minipage}[b]{0.4\textwidth}
  \centering
	\includegraphics[width=1.25\textwidth]{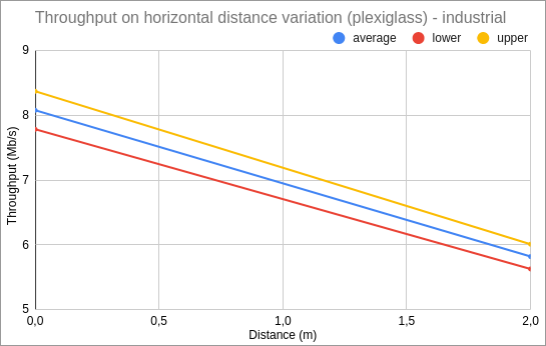}
	\caption{Throughput for horizontal distance variation in LoS case test when plexiglass acts as communication obstacle.}
	\label{fig:throughputreal}
  \end{minipage}
  \hfill
  \begin{minipage}[b]{0.45\textwidth}
  \centering
	\includegraphics[width=1.1\textwidth]{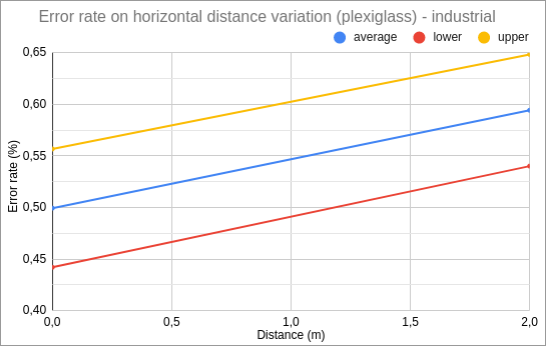}
	\caption{Error rate for horizontal distance variation in LoS case experiment when plexiglass is inserted as communication barrier.}
	\label{fig:errorratereal}
  \end{minipage}
\end{figure}

Figures \ref{fig:throughputreal} and \ref{fig:errorratereal} show the throughput and the packet error rate in the presence of transparent plexiglass barrier between the console and the lamp. The position  of the dongle   varies  horizontally from 0 to 2m. The throughput drops by about 25\% and the error rate increase of 20\%. This is consequence of the plexiglass introduction.

\subsection{Latency tests}
	
\subsubsection{"In vitro" tests}
\label{subsubsec:invitrorealtime}

\begin{table}[h]
	\begin{center}
		\begin{tabular}{|p{0.15\textwidth} | p{0.15088\textwidth} |p{0.15075\textwidth} | p{0.15075\textwidth} | p{0.1507\textwidth} |}
			\hline
			Scenario & Average(ms) & Lower(ms) & Upper(ms) & Peaks \\ \hline
			LoS  1 & 7,7124 & 7,7038 & 7,721 & 897 $10^{-7}$\\ \hline
			LoS  2 & 7,7516 & 7,4248 & 7,7608 & 901 $10^{-7}$\\ \hline
			NLoS 3 & 7,7582 & 7,7461 & 7,7703 & 900 $10^{-7}$\\ \hline
		\end{tabular}
	\end{center}
	\caption{RTTs test "in vitro".}
	\label{table:1}
\end{table}

We have tested the latency our 'in vitro' scenarios by using the following three scenarios: (Los 1) the height of the lamp is 2.5m while the horizontal distance $\Delta h$ is 0m (see Figure \ref{fig:vitro_los_vertical}); (Los 2) the height of the lamp is 2.5m while the horizontal distance $\Delta h$ is 1.2m (see Figure \ref{fig:vitro_los_vertical}); (NLos 1) the height of the lamp is 2.5m while the horizontal distance is 0m (see Figure \ref{fig:vitro_nlos_vertical}), the console has no line of sight with the OWC lamp and faces a surface at distance $\Delta h = 0.25$.  The latency results of these scenarios are shown in Table \ref{table:1}. We can see that for all scenarios the response time is always very low. This is consequence on the very small amount of data that is sent for each packet (512 byte) and its periodicity (every 100ms). We can also notice that in all scenarios that we have considered there is always a good throughput and a low PER (see Section \ref{ThroughputPER} for details). It is worth noting that the percentage of packets that exceeds the 30ms (that is the peaks) is extremely low in all scenarios. 
	
\subsubsection{Real-life environment test}
\begin{table}[h]
	\begin{center}
			\begin{tabular}{|p{0.1507\textwidth} | p{0.15088\textwidth} |p{0.15075\textwidth} | p{0.15075\textwidth} | p{0.1507\textwidth} |}
			\hline
			Scenario & Average(ms) & Lower(ms) & Upper(ms) & Peaks \\ \hline
			LoS 1 & 7,7337 & 7,7194 & 7,7479 & 908 $10^{-7}$\\ \hline
			LoS 2 & 7,7762 & 7,7619 & 7,7904 & 907 $10^{-7}$\\ \hline
		\end{tabular}
	\end{center}
	\caption{RTT tests in real-life environment.}
	\label{table:2}
\end{table}

We have tested the latency also for our industrial case study by using the following two scenarios: (Los 1) the height of the lamp is 5m while the horizontal distance $\Delta h$ is 2m (see Figure \ref{fig:errorratehorizontalnlosvitro15}); (Los 2) the height of the lamp is 5m while the horizontal distance $\Delta h$ is 2; a plexiglass between console and lamp is added (see Figure \ref{fig:reallife_plexi}). The latency results of these scenarios are shown in Table \ref{table:2}. We can see that in all scenarios the response time is always very low and the results obtained are equal to the 'in vitro' scenario.

\subsection{Discussion}
\label{chap:Discussion}
Our research questions were positively answered. The Light tecnonology has satisfactory performance in terms of latency, reliability and throughput when compared to wired and RF-based wireless communication
inside factories. While our experiments show that was possible to manoeuvre robots by using OWC, they also showed that WiFi could not be used because of various forms of interference. This lead to poor  reliability and  throughput.

Thanks to the high reliability of the light channel the dead-man button could be implemented. We recall that this is a switch that is activated or deactivated if the human operator becomes unable, such as through death, loss of consciousness, or being bodily  removed  from  control.   The  robot  is  allowed  to  move until  such  button  is  pressed,  once  it  is  released  the  robot  is prevented to move. The man dead button was impossible to implement over WiFi. 

\section{Conclusion and Future Work}\label{conclusion}
\label{chap:Conclusions}

We conducted various experiments to observe the suitability and efficiency of OWC based technology for antopomorphous robots control in industrial environments. The tests have shown overall satisfying results. In the "in vitro" tests, where the hardware was in the working range specified by the manufacturer, throughput and reliability tests have always shown acceptable results in LoS configuration. The real-life environment based tests presented satisfactory results in the assumed configurations. From the performance comparison between the "clear LoS" and the plexiglass tests, it's evident that a transparent obstacle is not that much of high concern for the technology, but it is highly preferable to have a clear LoS.
As for the real-timeness tests are concerned, they have shown very good results as the latency peaks are very few and the average latency is very low, hence making OWC a very good candidate for this kind of application.

We achieved all the proposed objectives, though for this kind of application there is the need for a specific implementation in order to make it usable when bigger distances are in play.

	\subsection{Future developments}
	In future research work, we aim to take measurements regarding light intensity using available hardware, so that can be sized a new hypothetical system which can be developed for simulation purposes. After the simulation phase, will be conducted an estimate of the hardware needed to build a prototype. Then will be built a prototype that implements the additional safety channel and a pointing system analogous to the one described in \ref{subsec:underwaternuclear}. Regarding the safety channel its communications will be processed separately from the motion's in order to respect industrial standards regarding safety and for the system to be as robust as possible, as this is the most critical feature in this application. We are also planning to implement  a pointing system. This is deemed as necessary in order to offer the operator the highest degree of movement while maintaining the optical link's quality as close as possible to the best conditions.
   
	
	\section{Acknowledgment}
	We thank the start-up Misco Valley for supporting all the experiments and providing the experimental hardware.
%
%
%
%
\bibliographystyle{elsarticle-num}
 \bibliography{cas-refs}

\begin{thebibliography}{10}
\expandafter\ifx\csname url\endcsname\relax
  \def\url#1{\texttt{#1}}\fi
\expandafter\ifx\csname urlprefix\endcsname\relax\def\urlprefix{URL }\fi
\expandafter\ifx\csname href\endcsname\relax
  \def\href#1#2{#2} \def\path#1{#1}\fi

\bibitem{Lu2017}
Y.~Lu, Industry 4.0: A survey on technologies, applications and open research
  issues, Journal of Industrial Information Integration 6 (2017) 1--10.
\newblock \href {https://doi.org/10.1016/j.jii.2017.04.005}
  {\path{doi:10.1016/j.jii.2017.04.005}}.

\bibitem{Wetzker2016}
U.~Wetzker, I.~Splitt, M.~Zimmerling, C.~A. Boano, K.~Romer, Troubleshooting
  wireless coexistence problems in the industrial internet of things, in: 2016
  {IEEE} Intl Conference on Computational Science and Engineering ({CSE}) and
  {IEEE} Intl Conference on Embedded and Ubiquitous Computing ({EUC}) and 15th
  Intl Symposium on Distributed Computing and Applications for Business
  Engineering ({DCABES}), {IEEE}, 2016.
\newblock \href {https://doi.org/10.1109/cse-euc-dcabes.2016.167}
  {\path{doi:10.1109/cse-euc-dcabes.2016.167}}.

\bibitem{martinek2019visible}
R.~Martinek, L.~Danys, R.~Jaros, Visible light communication system based on
  software defined radio: Performance study of intelligent transportation and
  indoor applications, Electronics 8~(4) (2019) 433.

\bibitem{biton2018improved}
C.~Biton, S.~Arnon, Improved multiple access resource allocation in visible
  light communication systems, Optics Communications 424 (2018) 98--102.

\bibitem{shao2015design}
S.~Shao, A.~Khreishah, M.~Ayyash, M.~B. Rahaim, H.~Elgala, V.~Jungnickel,
  D.~Schulz, T.~D. Little, J.~Hilt, R.~Freund, Design and analysis of a
  visible-light-communication enhanced wifi system, Journal of Optical
  Communications and Networking 7~(10) (2015) 960--973.

\bibitem{soner2019low}
B.~Soner, S.~C. Ergen, A low-swap, low-cost transceiver for physically secure
  uav communication with visible light, in: International Symposium on
  Innovative and Interdisciplinary Applications of Advanced Technologies,
  Springer, 2019, pp. 355--364.

\bibitem{ullah2020cognition}
Z.~Ullah, F.~Al-Turjman, L.~Mostarda, Cognition in uav-aided 5g and beyond
  communications: A survey, IEEE Transactions on Cognitive Communications and
  Networking (2020).

\bibitem{galisteo2019research}
A.~Galisteo, D.~Juara, D.~Giustiniano, Research in visible light communication
  systems with openvlc1. 3, in: 2019 IEEE 5th World Forum on Internet of Things
  (WF-IoT), IEEE, 2019, pp. 539--544.

\bibitem{ullah2020applications}
Z.~Ullah, F.~Al-Turjman, L.~Mostarda, R.~Gagliardi, Applications of artificial
  intelligence and machine learning in smart cities, Computer Communications
  (2020).

\bibitem{8732936}
F.~{Al-Turjman}, J.~P. {Lemayian}, S.~{Alturjman}, L.~{Mostarda}, Enhanced
  deployment strategy for the 5g drone-bs using artificial intelligence, IEEE
  Access 7 (2019) 75999--76008.

\bibitem{8624288}
F.~{Al-Turjman}, L.~{Mostarda}, E.~{Ever}, A.~{Darwish}, N.~{Shekh Khalil},
  Network experience scheduling and routing approach for big data transmission
  in the internet of things, IEEE Access 7 (2019) 14501--14512.

\bibitem{masini2018survey}
B.~M. Masini, A.~Bazzi, A.~Zanella, A survey on the roadmap to mandate on board
  connectivity and enable v2v-based vehicular sensor networks, Sensors 18~(7)
  (2018) 2207.

\bibitem{goto2016new}
Y.~Goto, I.~Takai, T.~Yamazato, H.~Okada, T.~Fujii, S.~Kawahito, S.~Arai,
  T.~Yendo, K.~Kamakura, A new automotive vlc system using optical
  communication image sensor, IEEE photonics journal 8~(3) (2016) 1--17.

\bibitem{raza2018social}
N.~Raza, S.~Jabbar, J.~Han, K.~Han, Social vehicle-to-everything (v2x)
  communication model for intelligent transportation systems based on 5g
  scenario, in: Proceedings of the 2nd International Conference on Future
  Networks and Distributed Systems, 2018, pp. 1--8.

\bibitem{siddiqi2016visible}
K.~Siddiqi, A.~Raza, S.~S. Muhammad, Visible light communication for v2v
  intelligent transport system, in: 2016 International Conference on Broadband
  Communications for Next Generation Networks and Multimedia Applications
  (CoBCom), IEEE, 2016, pp. 1--4.

\bibitem{bao2016visible}
Y.~Bao, Y.~Wang, J.~Yu, J.~Shen, A visible light communication based vehicle
  collision avoiding system, in: 2016 15th International Conference on Optical
  Communications and Networks (ICOCN), IEEE, 2016, pp. 1--3.

\bibitem{Dahri}
A.~B. Faisal Ahmed~Dahri, Hyder Bux~Mangrio, Experimental evaluation of
  intelligent transport system with vlc vehicle-to-vehicle communication,
  Wireless Personal Communications 106~(3) (2019) 1885–1896.

\bibitem{zhou2019common}
Y.~Zhou, X.~Zhu, F.~Hu, J.~Shi, F.~Wang, P.~Zou, J.~Liu, F.~Jiang, N.~Chi,
  Common-anode led on a si substrate for beyond 15 gbit/s underwater visible
  light communication, Photonics Research 7~(9) (2019) 1019--1029.

\bibitem{rust2012dual}
I.~C. Rust, H.~H. Asada, A dual-use visible light approach to integrated
  communication and localization of underwater robots with application to
  non-destructive nuclear reactor inspection, in: 2012 IEEE International
  Conference on Robotics and Automation, IEEE, 2012, pp. 2445--2450.

\bibitem{zhao2019preliminary}
W.~Zhao, M.~Kamezaki, K.~Yoshida, K.~Yama-guchi, M.~Konno, A.~Onuki, S.~Sugano,
  A preliminary experimental study on control technology of pipeline robots
  based on visible light communication, in: 2019 IEEE/SICE International
  Symposium on System Integration (SII), IEEE, 2019, pp. 22--27.

\bibitem{murai2012novel}
R.~Murai, T.~Sakai, H.~Kawano, Y.~Matsukawa, Y.~Kitano, Y.~Honda, K.~C.
  Campbell, A novel visible light communication system for enhanced control of
  autonomous delivery robots in a hospital, in: 2012 IEEE/SICE International
  Symposium on System Integration (SII), IEEE, 2012, pp. 510--516.

\bibitem{1429989}
R.~{Khalili}, K.~{Salamatian}, A new analytic approach to evaluation of packet
  error rate in wireless networks, in: 3rd Annual Communication Networks and
  Services Research Conference (CNSR'05), 2005, pp. 333--338.

\end{thebibliography}

\end{document}